\documentclass[letterpaper]{article} % DO NOT CHANGE THIS
\usepackage{aaai24}  % DO NOT CHANGE THIS
\usepackage{times}  % DO NOT CHANGE THIS
\usepackage{helvet}  % DO NOT CHANGE THIS
\usepackage{courier}  % DO NOT CHANGE THIS
\usepackage[hyphens]{url}  % DO NOT CHANGE THIS
\usepackage{graphicx} % DO NOT CHANGE THIS
\usepackage{natbib}  % DO NOT CHANGE THIS AND DO NOT ADD ANY OPTIONS TO IT
\usepackage{caption} % DO NOT CHANGE THIS AND DO NOT ADD ANY OPTIONS TO IT
\usepackage{algorithm}
\usepackage{algorithmic}
\usepackage{subcaption}

\usepackage{amsmath}
\usepackage{amssymb}
\usepackage{booktabs}
\usepackage{multirow}

\usepackage{newfloat}
\usepackage{listings}
\DeclareCaptionStyle{ruled}{labelfont=normalfont,labelsep=colon,strut=off} % DO NOT CHANGE THIS
\floatstyle{ruled}
\newfloat{listing}{tb}{lst}{}
\floatname{listing}{Listing}
\title{MFABA: A More Faithful and Accelerated Boundary-based Attribution Method for Deep Neural Networks}
\author {
    Zhiyu Zhu\textsuperscript{\rm 1},
    Huaming Chen\textsuperscript{\rm 1},
    Jiayu Zhang\textsuperscript{\rm 2},
    Xinyi Wang\textsuperscript{\rm 3},
    Zhibo Jin\textsuperscript{\rm 1},
    Minhui Xue\textsuperscript{\rm 4}\\
    Dongxiao Zhu\textsuperscript{\rm 5},
    Kim-Kwang Raymond Choo\textsuperscript{\rm 6}
}
\affiliations {
    \textsuperscript{\rm 1}The University of Sydney,
    \textsuperscript{\rm 2}SuZhouYierqi,
    \textsuperscript{\rm 3}University of Malaya,
    \textsuperscript{\rm 4}CSIRO's Data61\\
    \textsuperscript{\rm 5}Wayne State University,
    \textsuperscript{\rm 6}University of Texas at San Antonio
}

\begin{document}

\maketitle

\begin{abstract}
To better understand the output of deep neural networks (DNN), attribution based methods have been an important approach for model interpretability, which assign a score for each input dimension to indicate its importance towards the model outcome. Notably, the attribution methods use the axioms of sensitivity and implementation invariance to ensure the validity and reliability of attribution results. Yet, the existing attribution methods present challenges for effective interpretation and efficient computation. In this work, we introduce MFABA, an attribution algorithm that adheres to axioms, as a novel method for interpreting DNN. Additionally, we provide the theoretical proof and in-depth analysis for MFABA algorithm, and conduct a large scale experiment. The results demonstrate its superiority by achieving over 101.5142 times faster speed than the state-of-the-art attribution algorithms. The effectiveness of MFABA is thoroughly evaluated through the statistical analysis in comparison to other methods, and the full implementation package is open-source at: \url{https://github.com/LMBTough/MFABA}.
\end{abstract}

\section{Introduction}
	Deep learning (DL) has shown prominent performance in various areas of computing tasks, such as image classification \cite{li2022research}, semantic segmentation \cite{mo2022review}, object detection \cite{zaidi2022survey}, and text classification \cite{minaee2021deep}. A wide range of applications in practice have demonstrated its effectiveness, unfortunately without much contextual explanation of the decision process. This has led to a severe crisis towards the trustworthiness of DL models given the facts of poor interpretability of results, intractability of model errors, and the difficulty in tracing model behaviours \cite{janik2019interpreting}. It remains challenging for researchers to obtain a better understanding of complicated models, especially those based on multiple layers and designed for nonlinear learning.
	
	Recently, the attribution methods have been proposed as one of the most promising means to solve this problem, which can find the causal relationship between the inputs and outputs. In general, there are two fundamental axioms proposed in Integrated Gradients (IG) \cite{sundararajan2017axiomatic}: Sensitivity, and Implementation Invariance. Sensitivity requires good capability of differing the feature and prediction for every input, which ensures that the input information can be correctly attributed for the predictions. In addition, a method that satisfies Implementation Invariance defines that two neural networks with the same input and output values are functionally equivalent regardless the implementation details. The attribution method should retain the same results given two identical networks. 
	
 Different from IG using integration to calculate the contribution, Expected Gradient (EG) method ~\cite{erion2021improving} introduces prior knowledge and uses it as a prior probability distribution for feature attribution, proposing expectation gradients to calculate the importance of an input feature to the output. Boundary-based Integrated Gradient (BIG) method is one of the first methods that uses adversarial attacks with a linear attribution path to identify appropriate decision boundaries for interpretation~\cite{wang2021robust}. However, the overall performance is limited due to the linear attribution path of IG algorithm~\cite{jin2023danaa}, and BIG requires much more time for computation. 

    Adversarial Gradient Integration (AGI) method ~\cite{pan2021explaining} exploits the gradient information of adversarial examples to compute the contribution of all input features by integrating the gradient along the non-linear path with the steepest ascent. While AGI does not depend on the choice of reference points in IG, it employs targeted adversarial attacks to discern optimal decision boundaries. Consequently, AGI incurs significant computational costs, particularly in complex tasks and models where extensive gradient integrations are required. The interpretation performance is further hindered if noise or outliers exist in the input data.

    To address the noise pixels generated by IG in regions where the prediction category is irrelevant, the Guided Integrated Gradients (GIG) method~\cite{kapishnikov2021guided} sets all contributions in the region specified in the feedforward process to zero by guided gradients and only the target region of the network output needs to be considered. While only effective for limited datasets, GIG has a high demand on computational resources and time cost.
 
	Thus, current attribution algorithms is yet to provide accurate attribution results 
%(That is, the alteration in the loss function is equivalent to the summation of attribution outcomes) 
and require significant computational cost. To tackle these issues, in this work, we propose a new attribution method, called More Faithful and Accelerated Boundary-based Attribution method (MFABA) which attempts to exploits part of the adversarial attack nature for a faster and more effective attribution. MFABA proposes a novel idea based on the second-order Taylor expansion of the loss function in addition to IG, which attribution results demonstrate a more faithful performance. We also investigate the attack from the attribution method to explore the limitations for the linear paths. 
	
	In summary, the contributions of this paper are as follows: 
 \begin{itemize}
     \item A novel attribution method MFABA is proposed which demonstrate a state-of-the-art performance; 
     \item A detailed derivation of attribution validity and an axiomatic are presented for MFABA; 
     \item The definition of attack in MFABA is provided, and we substantiate the superiority of MFABA through evidence;
     \item The replication package of MFABA is released.
 \end{itemize}

\section{Background}
	\subsection{From Attribution to Interpretability}
	The interpretability of DL model refers to the ability of explain the predictions and decisions made by the model in a way that human can understand \cite{zhang2018visual}. For the attribution method in DNN, an exact one-to-one correspondence between the input and output should be provided \cite{ancona2017towards}. Thus, attribution method is considered an exclusive set of methods providing the interpretability for DL models. Other general interpretable algorithms may not meet the axiomatic requirements of Sensitivity and Implementation Invariance \cite{sundararajan2017axiomatic}. 
	
	For most visual related tasks, attribution methods target on the corresponding information for the feature and prediction. It has thus been challenging for comprehensive evaluation with human intuition, which is more applicable with other interpretation approaches, such as Grad-CAM \cite{selvaraju2017grad} and Score-CAM \cite{wang2020score}. These methods are gradient-based methods, among which Grad-CAM faces different challenges like gradient saturation \cite{ramaswamy2020ablation}, gradient disappearance \cite{zhang2019improved}, and low performance for both the coarse-grained heat map generated at the deep level and the fine-grained heat map generated at the superficial level \cite{choi2020interpreting}. Similarly, Score-CAM aims to bypass the gradients reliance and can obtain the weights of each activation map through forward propagation of scores on target classes. Eventually a linear combination of the weights and activation map will be obtained. However, both methods fail to provide an accurate attribution for the features. Only the intermediate layers of the networks are interpreted, yielding intuitively interpretable results at the corresponding layer. Yet, the methods do not satisfy the two fundamental axioms of the attribution methods. 

    Saliency Map (SM)~\cite{simonyan2013deep,patra2020incremental} is one earliest attribution method aiming at the visualisation of the particular features related to model outputs. In details, the partial derivative of the loss function $\partial f(x)$ is leveraged to calculate the degree of importance of $x$. SM suffers from gradient saturation and does not satisfy the axiom of Sensitivity. For example, in a simple neural network $f(x)=1-\operatorname{ReLU}(1-x)=\left\{\begin{array}{l}x, x<1 \\\ 1, x \geq 1\end{array}\right.$, the attribution result could be 0 at $x=0$ or $x=2$. However, it will be none for $\partial f(x)$. To address this issue, Integrated Gradient (IG)~\cite{sundararajan2017axiomatic} integrates the gradients over different paths to obtain the degree of contribution of the non-zero gradient in the non-saturated region. However, it suffers from: 
	(1) poor results due to the choice of baseline may lead to significant deviations~\cite{pan2021explaining}. It is not feasible to find an appropriate baseline for various tasks.
	(2) expensive computational process, which requires multiple rounds of propagation to obtain approximate integration results.
	(3) ineffective representation of sample transformation path from the selected gradient path, presenting as an attack conflict issue.

    Other methods, like FullGrad method~\cite{srinivas2019full}, aim to use local gradient information to interpret DNN internal structure. Distilled Gradient Aggregation (DGA) method~\cite{jeon2022distilled} considers the linear regions of decision boundaries based on intermediate local attribution for a sequence of meaningful baseline points. LIME algorithm~\cite{ribeiro2016should} amalgamates approximation techniques with weighted sampling methods to construct a local model to generate interpretable predictions from the model classifier. Shapley Additive Explanations (SHAP) algorithm~\cite{lundberg2017unified} computes feature contribution to the prediction outcome using Shapley values and subsequently ranks their importance, thereby achieving both local and global interpretation of the model. However, these methods still face the problem of low accuracy and slow attribution speed. Furthermore, LIME and SHAP tend to emphasize axioms that ensure `local faithfulness' or `local accuracy' of explanations, deviating from strict and complete attributions based on sensitivity and implementation invariance axioms.

	\subsection{Gradient-based Adversarial Attack}
	\subsubsection{Adversarial Attack Algorithms}
	Adversarial attack aims to find the minimum perturbation of input to deviate the output result. In our work, the baseline for finding the attribution and the computational process for optimizing the attribution algorithm are the same. We firstly discuss their relationships and the differences.
	\begin{equation}
		x_j=x_{j-1}+\eta \operatorname{sign}\left(\nabla_x L(\theta, x, y)\right)
		\label{equ2}
	\end{equation}
	\begin{equation}
		x^{t+1}=\Pi_{x+\mathcal{S}}\left(x^t+\alpha \operatorname{sign}\left(\nabla_x L(\theta, x, y)\right)\right)
		\label{equ3}
	\end{equation}
	Eq.~\ref{equ2} represents the attack process for FGSM~\cite{goodfellow2014explaining}, where a clamp gradient ascent method is used to perform the attack. For I-FGSM~\cite{kurakin2018adversarial}, multi rounds of FGSM attacks are conducted to identify the optimal perturbation direction. Eq.~\ref{equ3} represents the attack process for PGD, where the distance of the attack is limited by mapping the attacks to the circular space of the data, thereby making the attack result as close as possible to the actual result.
	
	\subsubsection{Definition of Successful Attack}
	In the classification task, $x_0$ belonging to the target category of $C$ and an adversarial sample $x_n$ producing an output category different from $C$ denotes a successful attack, subject to $L_{P}$ norm for $x_0$ and $x_n$ is less than $\varepsilon$. Attack samples with larger deviations will be considered as failed attack, as shown in Eq.~\ref{equ4}.	
        \begin{equation}
		m\left(x_0\right)=C \text { and } m\left(x_n\right) \neq C \text { and } \left\|x_0-x_n\right\|_p<\varepsilon
		\label{equ4}
	\end{equation}

\section{Method}
	In this section, we first present the derivation of gradient ascent, with the axiomatic proof for our attribution method. Secondly, we discuss the different methods of `sharp' and `smooth' gradient ascent for MFABA.
	\subsection{Theoretical Derivation}
	\subsubsection{Gradient Ascent Method}
        Inspired from BIG and AGI, we consider that gradient ascent of the loss function can push the adversarial samples across the decision boundary of the model. The model's inference response on these adversarial samples is the key information for interpretation. Following Eq.~\ref{equ5}-\ref{equ7} present the gradient ascent process using first-order Taylor expansion. $L$ is the loss function.
	\begin{equation}
		L\left(x_{j}+\alpha d\right)=L\left(x_{j}\right)+\alpha g_{j}^T d+\varepsilon
        \label{equ5}
	\end{equation}
	\begin{equation}
		L\left(x_{j}+\alpha d\right)>L\left(x_{j}\right) \quad s.t. \quad g_{j}^T\cdot d>0
	\label{equ7}
        \end{equation}
	In which, $g_{j}=\frac{\partial L\left(x_{j}\right)}{\partial x_{j}}$, $d$ is an update direction vector with the same dimension as $x_{j}$. We can get $L\left(x_{j}+\alpha d\right)>L\left(x_{j}\right)$ if $g_{j}^T\cdot d>0$, where $\cdot$ represents dot product. Then we use $x_{j+1}=x_{j}+\alpha d$ or $x_{j+1}=x_{j}+sign(\alpha d)$ to update the adversarial sample. The $sign$ function here meets the decoupling requirements of adversarial attacks. To achieve this, the scalar learning rate $\alpha$ for the gradient ascent process will be minimum. 
 
    Since the first-order Taylor expansion only takes into account the gradient (first derivative) but ignores curvature (second derivative) of the loss function, it offers a less comprehensive source of information compared to the second-order Taylor expansion.To more accurately depict the local behavior of the model in the vicinity of a given input point, particularly accounting for the non-linear impact of features, we consider that the corresponding function $L$ can be transformed with the second-order Taylor expansion at the point $x_j$ in our MFABA. Next is the derivation of how to obtain the attribution of each model input in MFABA.

	\subsubsection{MFABA Mathematical Derivation} Here, we list second-order Taylor expansion of Eq.~\ref{equ5} during gradient ascent:
 
    \begin{align}
        L\left(x_j\right) &= L\left(x_{j-1}\right) + \frac{\partial L\left(x_{j-1}\right)}{\partial x_{j-1}}\left(x_j-x_{j-1}\right) \nonumber \\
        &\quad + \frac{1}{2}\frac{\partial^2L(x_{j-1})}{\partial x_{j-1}^{2}}(x_{j}-x_{j-1})^{2} + \varepsilon 
        \label{equ9}
    \end{align}
    Eq.~\ref{equ9} indicates that second-order Taylor expansion can be performed when $x_j$ and $x_{j+1}$ are close.

    \begin{align}
        \sum_{j=1}^n L\left(x_j\right) &= \sum_{j=0}^{n-1} L\left(x_j\right) + \sum_{j=0}^{n-1} \frac{\partial L\left(x_j\right)}{\partial x_j}\left(x_{j+1}-x_j\right) \nonumber \\
        &\phantom{=}+ \sum_{j=0}^{n-1} \frac{1}{2}\frac{\partial^2L(x_{j})}{\partial x_{j}^{2}}(x_{j+1}-x_{j})^{2} 
        \label{equ10}
    \end{align}

     \begin{equation}
        \begin{split}
            L\left(x_n\right)-L\left(x_0\right)&= \sum_{j=0}^{n-1} \left(\frac{\partial L\left(x_j\right)}{\partial x_j}\left(x_{j+1}-x_j\right) \right. \\
            &\qquad \left. +\frac{1}{2}\frac{\partial^2L(x_{j})}{\partial x_{j}^{2}}(x_{j+1}-x_{j})^{2}\right)
        \end{split}
        \label{equ11}
    \end{equation}
    In Eq.~\ref{equ10} and Eq.~\ref{equ11}, we derive the approximate derivation relationship from Eq.~\ref{equ9}, and $\varepsilon$ is omitted.

        \begin{equation}
        \resizebox{0.42\textwidth}{!}{$
            \begin{aligned}
                \frac{\partial^2L(x_{j})}{\partial x_{j}^{2}}(x_{j+1}-x_{j})^{2}
                & = (\frac{\partial L(x_{j+1})}{\partial x_{j+1}}-\frac{\partial L(x_{j})}{\partial x_{j}})(x_{j+1}-x_{j})=\bigtriangleup x^{T}H\bigtriangleup x \\
                \phantom{{}={}} &= \bigtriangleup x^{T}\cdot \phantom{{}={}} \begin{bmatrix}
                     h_{11}\cdot \bigtriangleup x^{1}+h_{12}\cdot \bigtriangleup x^{2}+...+h_{1n}\cdot \bigtriangleup x^{n} \\
                     \dots \\
                     h_{n1}\cdot \bigtriangleup x^{1}+h_{n2}\cdot \bigtriangleup x^{2}+...+h_{nn}\cdot \bigtriangleup x^{n}
                \end{bmatrix}
            \end{aligned}
        $}
    \label{equsecond}
    \end{equation}
    
    In Eq.~\ref{equsecond}, we use the Hessian matrix $H$ to calculate the second-order derivative part in the Taylor expansion. 

    We replace the second derivative in Eq.~\ref{equ11} with the Hessian matrix proposed in Eq.~\ref{equsecond}, and finally Eq.~\ref{equfinal} is as follows:
    \begin{equation}
        \scalebox{0.78}{$
            \begin{aligned}
                L\left(x_n\right)-L\left(x_0\right)
                &= \sum_{j=0}^{n-1} \Bigg(\frac{\partial L\left(x_j\right)}{\partial x_j}\left(x_{j+1}-x_j\right) \\
                &\hphantom{{}={}} +\frac{1}{2}\left(\frac{\partial L(x_{j+1})}{\partial x_{j+1}}-\frac{\partial L(x_{j})}{\partial x_{j}}\right)(x_{j+1}-x_{j})\Bigg)\\
                &= \sum_{i=0}^p \Bigg(\sum_{j=0}^{n-1} \frac{\frac{\partial L\left(x_j\right)}{\partial x_j^i}+\frac{\partial L\left(x_{j+1}\right)}{\partial x_{j+1}^i}}{2}\left(x_{j+1}^i-x_j^i\right)\Bigg)
            \end{aligned}
        $}
    \label{equfinal}
    \end{equation}
	where $p$ indicates the size of the input dimension. And $x_j$ denotes the sample at the $j$-th gradient ascent, $x_0$ and $x_n$ are the original and adversarial sample, respectively. Eq.~\ref{equfinal} indicates that the difference between $L\left(x_0\right)$ and $L\left(x_n\right)$ can be seen as the sum of the attribution at each position. In other words, whenever the output values change between them, attribution of non-zero outcomes will be calculated, which meets the axiom of Sensitivity. By iteratively performing gradient ascent and computing the sum of attributions at each position, we can observe how these perturbed features influence the decision-making behavior of the model. Thus, for $\sum_{j=0}^{n-1} \frac{\frac{\partial L\left(x_j\right)}{\partial x_j^i}+\frac{\partial L\left(x_{j+1}\right)}{\partial x_{j+1}^i}}{2}\left(x_{j+1}^i-x_j^i\right)$, it can be seen as an attribution on the $i$-dimensional input.

	\subsection{Axiomatic Proof}
    Following we discuss the axioms satisfied in MFABA.
	\subsubsection{Definition of Sensitivity}An attribution method satisfies Sensitivity(a) if for every input and baseline that differ in one feature but have different predictions then the differing feature should be given a non-zero attribution.
	
	According to Eq.~\ref{equ11}, the sum of all imputations is $L\left(x_n\right)-L\left(x_0\right)$. Non-zero imputation results are always calculated when $x_0$ and $x_n$ lead to a change in $L$. Therefore our method follows the axiom.
 
        \subsubsection{Definition of Implementation Invariance}A method that satisfies Implementation Invariance should ensure that two neural network attributions with the same input and output values are consistent.
	It is clear that the computational processes in MFABA follow the chain rule of gradients, which meets the definition of implementation invariance~\cite{sundararajan2017axiomatic}.
 
	\subsection{Attribution Method in MFABA}
    Based on Eq.~\ref{equfinal}, the attribution corresponding to the $i$-dimensional input can be expressed as follows:

    \begin{equation}
\resizebox{0.4\textwidth}{!}{$
\begin{aligned}
MFABA(x^i) = &\sum_{j=0}^{n-1}\frac{1}{2}\left(\frac{\partial L(x_j)}{\partial x_j^i}+\frac{\partial L(x_{j+1})}{\partial x_{j+1}^i}\right) \\
&\cdot (x_{j+1}^i-x_j^i)
\end{aligned}
$}
\label{equ13}
\end{equation}

    In order to achieve best attribution results, the Taylor expansion needs to ensure that $x_{j+1}^i$ is close to $x_j^i$. We further use the approximation $\left(\frac{\partial L(x_{j+1})}{\partial x_{j+1}^{i}}-\frac{\partial L(x_{j})}{\partial x_{j}^{i}}\right)$ to replace the Hessian matrix, as it may be computationally expensive. Thus, little additional computational time is needed since there is no additional forward and backward propagation.
	
	The adversarial sampling will stop when the decision boundary is found (e.g., a category shift in a classification problem) to avoid potential bias in a sample. We have pratically set a maximum $n$ for gradient ascending step to mitigate additional computing costs in the absence of identified decision boundaries. Meanwhile, the function $L$ is broadly explored in experiments with comparison with BIG and IG, in which neural network output value is selected for attribution.
    However, we observed that a negative attribution may be generated. Suppose we have a toy sample for a three-classification task, the adversarial attack attempts to alter the output value of the model from [0.4, 0.5, 0.55] to [0.5, 0.65, 0.6]. Here 0.4, 0.5 and 0.55 represent the output values of class A, B and C respectively. It is obvious that after perturbation, the final output class of the model changes from $C$ to $B$. But the confidence value for class $C$ is actually increased, resulting in attribution errors when other models like IG and BIG use the probability results obtained before softmax function as the model output. We observe that the probability results obtained via softmax function can highlight the correct reduced probabilities for classification, which helps to mitigate the attribution issue. Thus, in MFABA, softmax output is used to attribute the category values. %A detailed comparison is presented in the experiment.

\subsection{Sharp and Smooth Gradient Ascent Methods}
For MFABA, two gradient ascent methods are applied namely sharp and smooth gradient ascent methods in Eq.~\ref{equ14}. Smooth method truncates the gradients, causing a relatively weak sample gradient to traverse a same distance as a strong sample gradient. For example, a pixel with 0.01 gradient will change by the same magnitude as a pixel with 0.71 gradient. Sharp method maximises the directionality of the preserved gradient in favour of the more dominant gradient information, and the attribution results in the sharpest information.
    \begin{equation}
    \begin{aligned}
		smooth (grad)=sign(grad) \\
		sharp (grad) = \frac{\text { grad }}{\| \text { grad } \|_{2}}
    \label{equ14}
    \end{aligned}
    \end{equation}

\subsection{The role of the $sign$ function}
	In MFABA, the gradient ascent method utilizes the adversarial attack to find samples and the equivalent gradient direction. Normally, the gradient is calculated as $\frac{\partial F_i}{\partial x_i}$. While the adversarial attack usually chooses a loss function as the objective, the gradient will be $\frac{\partial L}{\partial x}=\frac{\partial L}{\partial F_i} \frac{\partial F_i}{\partial x}=-\frac{1}{F_i}\frac{\partial F_i}{\partial x}$, where $\frac{1}{F_i}$ only affects the vector norm not the gradient direction, and can be interpreted as an equivalence relationship with the sharp and smooth methods. In classical adversarial attack task, $sign$ function (Eq.~\ref{equ14}) is required to prevent biased training towards the direction of larger weight \cite{goodfellow2022explaining}, preventing meaningless input changes for attribution results.

\section{In-depth analysis}

	\subsection{Analysis of MFABA Efficiency}
	We evaluate the method based on the inference speed and the number of forward and backward propagations in a unified GPU environment. The corresponding gradient information of $\frac{\partial L\left(x_j\right)}{\partial x_j}$ is kept during the gradient ascending process, which avoid recomputing for the subsequent steps. Typically, it takes 3-10 steps to find an adversarial sample. %which saves lots of computational tasks in forward and backward propagation. 
    In comparison with IG which also meets the axioms, IG requires 30-200 rounds of forward propagation between $x_0$ and $x_n$ while our algorithm only requires 3-10 rounds of forward and backward propagation. 
	
       \begin{figure}[htbp]
        \centering
        \includegraphics[width=\linewidth]{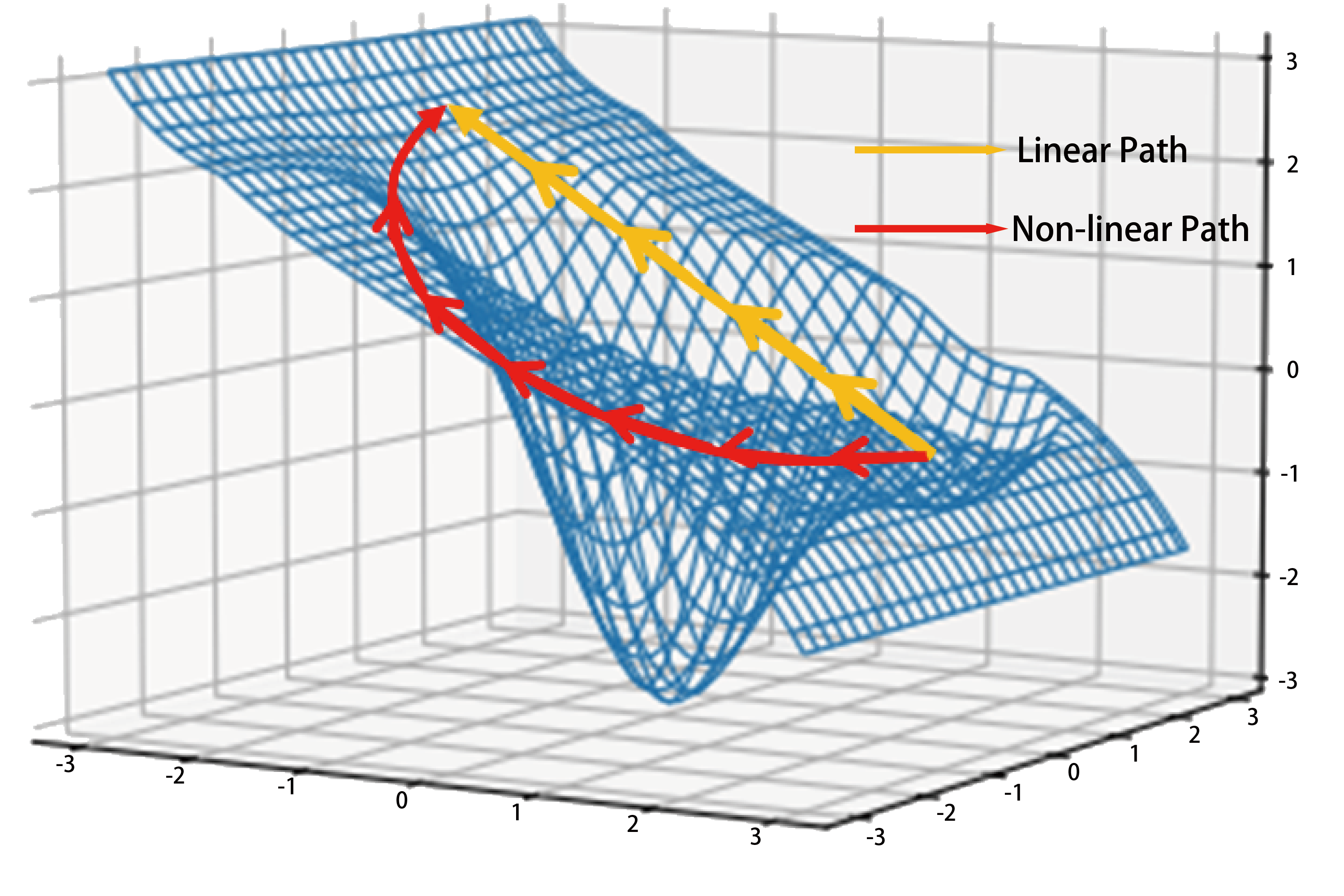}
        \caption{Linear and non-linear path of the direction for adversarial attack and baseline samples}
        \label{img:attack_direction}
        
    \end{figure}
    
	Different from the pairwise attack approach used by BIG, MFABA algorithm does not use boundary search, instead using gradient ascent method to identify the decision boundary with adversarial samples. With Eq.~\ref{equ13}, the samples obtained from each iteration are relatively close to each other, leading to high-quality samples for decision boundary. Other works may specify a directed adversarial sampling attack to find the attribution results, such as~\cite{pan2021explaining}. MFABA does not specific a direction of the adversarial attack to find a sample. The reason is that, for example, given a correct label of $A$, the adjacent decision boundary is defined as $B$. However, $C$ is closer to $B$. In this case, a directed adversarial attack to $C$ will not help to find the decision boundary for $A$ correctly. Also, directly using the adversarial attack to find decision boundary would result in extensive gradient computation and adversarial sampling time. In MFABA, we avoid this by preserving the gradient graph for forward and backward propagation.

        \subsection{Definition of Aggressiveness}
	\textbf{Definition:} When $x'$ satisfies the attack
	\begin{equation}
		L\left(x^{\prime}\right)>L(x)
	\end{equation}
	\begin{equation}
		\left\|x-x^{\prime}\right\|_p<\varepsilon
	\end{equation}
	If a lower value of loss function $L$ indicates a better performance for neural network, hereby we define the aggressiveness for $x'$, meaning that $x'$ is an sample with aggressiveness for $L$ function. In other way, the sample of $x'$ not subject to the equations is called a non-aggressive sample.

\begin{figure}[htbp]
	\begin{center}
	   \centerline{\includegraphics[width=.9\columnwidth]{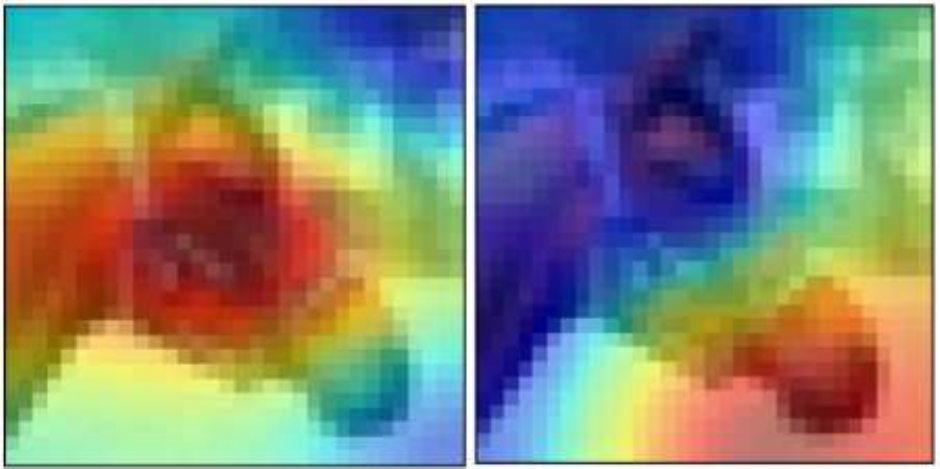}}
		\caption{Comparison of heatmap without (left) and with (right) non-aggressive samples}
		\label{img:ablation_aggressive}
	\end{center}
\end{figure}

    In Figure~\ref{img:attack_direction}, the red line denotes the direction of attack while yellow line represents the baseline direction, such as the linear path in BIG. As shown in the diagram, there may exist many non-aggressive samples. BIG algorithm needs to compute $\int_{0}^{1}\frac{\partial f((x-x_{b})t+x_{b})}{\partial x}dt$ in the process of attribution, changing $x_{b}$ to $x$ will result in many samples that do not have non-aggressive samples entering the computational process, and the process of non-aggressive sample computation will affect the correct attribution of the features. As shown in Figure.~\ref{img:ablation_aggressive}, in the process of MFABA calculation, we remove the non-aggressive sample features and visualize the aggressive or non-aggressive sample integration process separately. More details can be found in Appendix folder in GitHub link. 
    We find that all non-aggressive samples deviated from the key features have critical impacts on the results. We will further investigate this in the next section.

	\subsection{Comparison with Other State-of-the-art Methods}
    In this section, we will derive a linear approximate version of MFABA, and compare it with other methods, including IG and BIG. In BIG \cite{wang2021robust}, Eq.~\ref{equ17} shows that the computation can be seen as calculating the difference between $x$ and $x'$ and the definite integral of the computed gradient in the linear path, respectively.

{\small
\begin{small}
	\begin{equation}
		g_{BIG}(x;x_{b})=(x-x_{b})\int_{0}^{1}\frac{\partial f((x-x_{b})t+x_{b})}{\partial x}dt
        \label{equ17}
	\end{equation}
\end{small}
}

    Based on the discussion in Definition of Aggressiveness, we notice that the linear path may not be optimal. In MFABA, the non-aggressive samples are removed from the linear path. Regarding the overall computational tasks, the linear path requires more time to obtain the corresponding gradient information, at least 30-200 rounds in BIG for forward and backward propagation. 
    At this point, we consider mapping all the attack samples generated in the adversarial attack to the linear distance, and approximating the integration result using IG algorithm.

	\begin{equation}
		t_j=\frac{\sum_{i=1}^j\left\|x_i-x_{i-1}\right\|_p}{\sum_{k=1}^n\left\|x_k-x_{k-1}\right\|_p}
        \label{equ18}
	\end{equation}
	\begin{equation}
		t_j=\frac{\left(x_j-x_0\right) \cdot \cos <x_j-x_0, x_n-x_0>}{x_n-x_0}
        \label{equ19}
	\end{equation}

          \begin{figure}[htbp]
	\begin{center}
	   \centerline{\includegraphics[width=\columnwidth]{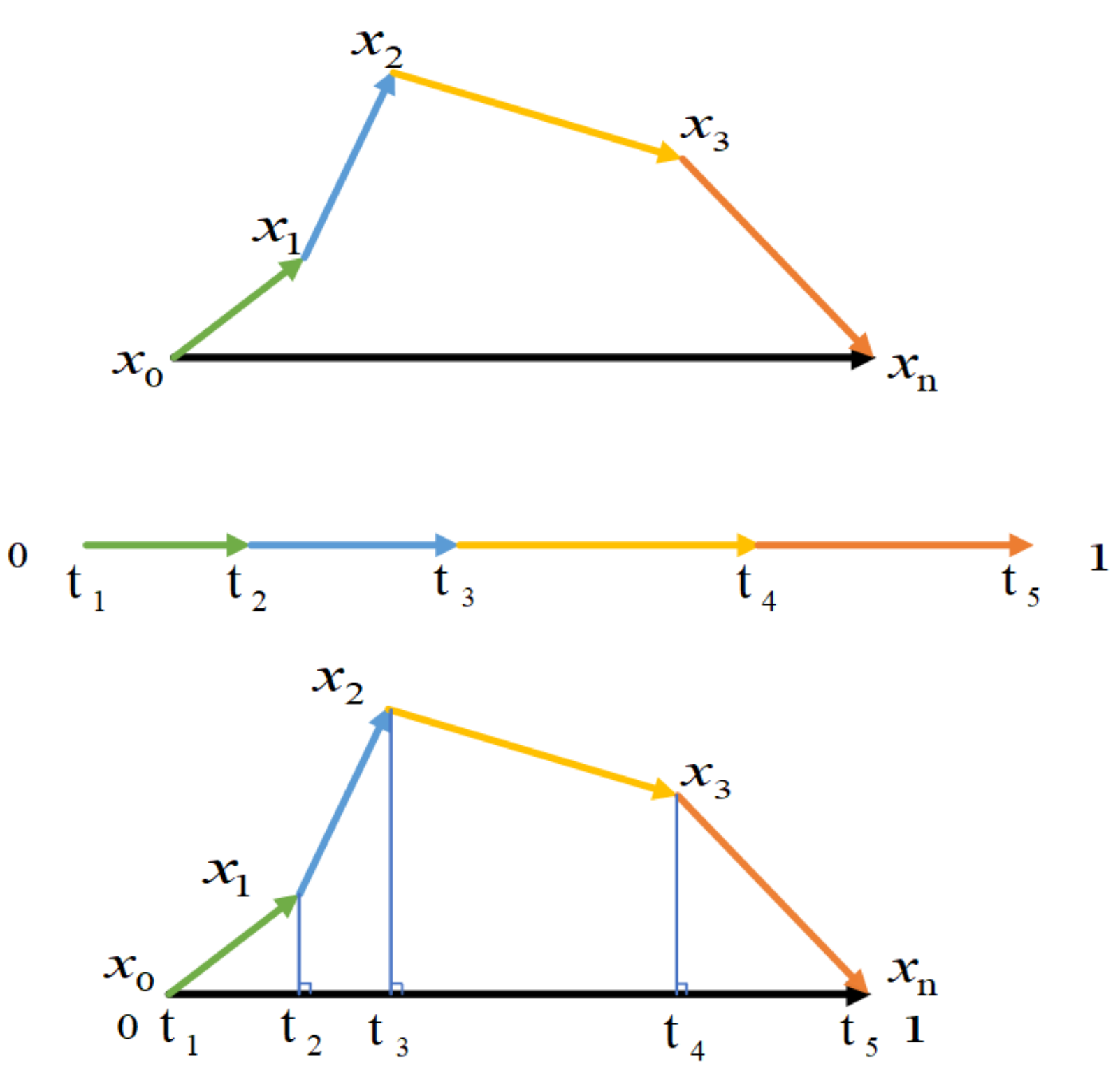}}
		\caption{Two approximate algorithms of MFABA. The top graph represents the MFABA-norm algorithm, and the bottom graph represents the MFABA-cosine algorithm}
		\label{img:approximations}
	\end{center}
\end{figure}
 
	Eq.~\ref{equ18} is MFABA-norm and Eq.~\ref{equ19} is MFABA-cosine method. Figure.~\ref{img:approximations} illustrates the two approximate algorithms of MFABA for Eq.~\ref{equ18} and~\ref{equ19}, respectively. Eq.~\ref{equ18} treats all the Lp-parametric distances of the motion trajectories $x_0$-$x_n$ as 1, and the position of $j$-th sample between 0 and 1 can be regarded as the relative distance of the position traveled by the $x_j$ sample. The second approach in Eq.~\ref{equ19} maps the relative positions of $x_j$ and $x_0$ onto the vector $x_0$-$x_n$, which also reaches the objective of obtaining the relative distance. We have included more visualization results of MFABA-norm and MFABA-cosine in the Appendix folder.

\section{Evaluation}
In this section, we provide the experiment design details and the experimental results to answer following questions: 1) Can MFABA provide enhanced and faithful interpretations of the model results in comparison with other state-of-the-art methods? 2) How much improvement has been achieved for MFABA in terms of the computational efficiency?
	\subsection{Experiment Setup}
	To fairly evaluate MFABA and other state-of-the-art methods, we have implemented the experiments with publicly available and widely used model architecture including Resnet50 and EfficientNet. The empirical experiments are conducted against the CIFAR10~\cite{ krizhevsky2010cifar}, CIFAR100 and ImageNet~\cite{russakovsky2015imagenet} datasets. It is acknowledged that the sizes of these datasets sizes exceed ten thousand. For each method, 50 gradient ascending steps are set as the maximum steps for attacking, and the learning rate is set to 0.01.
%    The evaluation approach employed in this paper is to reveal all the generic reproducible experimental nature of the code and to perform extensive visualisations to assess specific validity.
	
    \subsection{Empirical Evaluation}
\begin{figure}[htbp]
    \centering
    
    \begin{subfigure}{\linewidth}
        \centering
        \includegraphics[width=\linewidth]{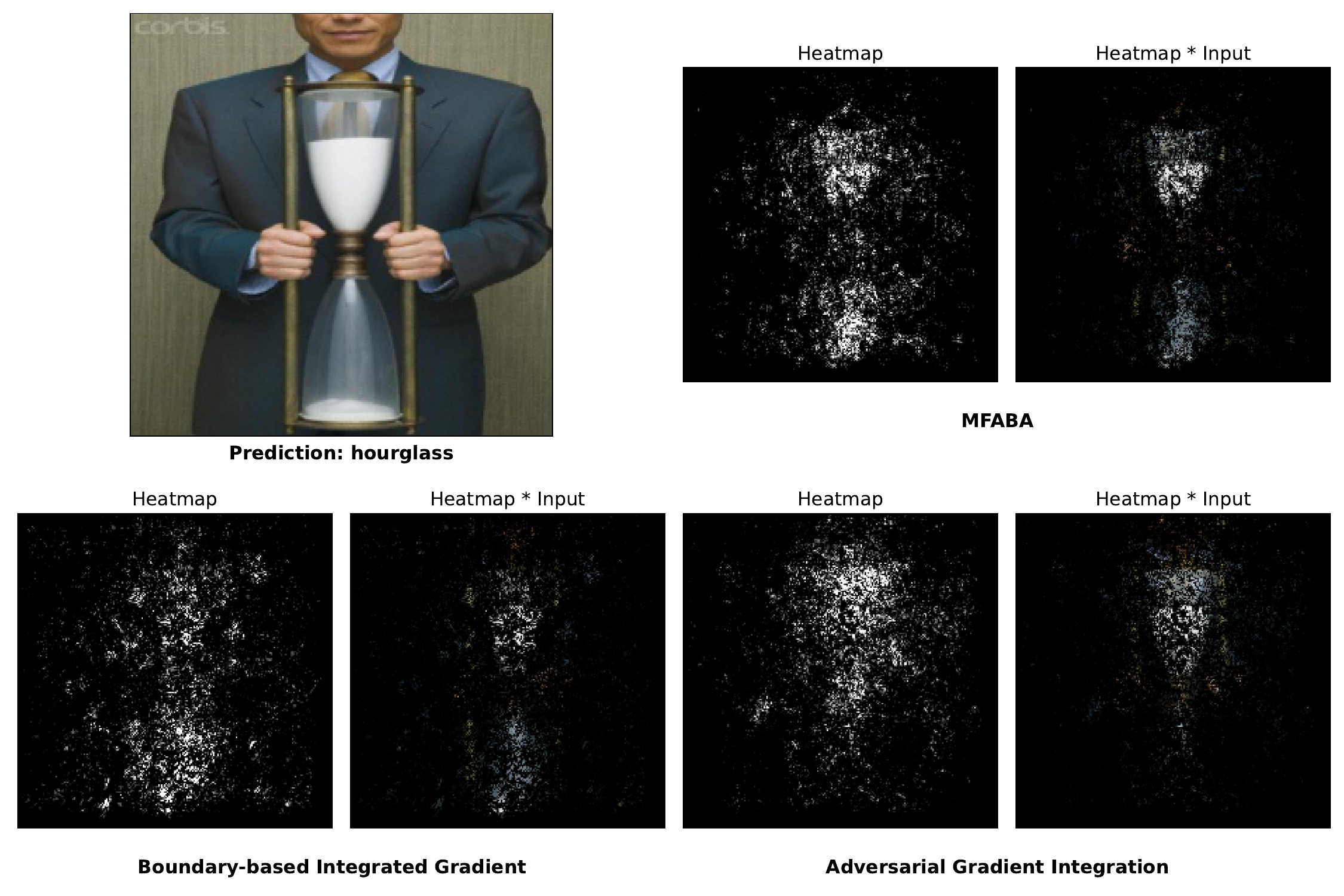}
        \label{subfig:empirical_evaluation}
    \end{subfigure}
    
    \begin{subfigure}{\linewidth}
        \centering
        \includegraphics[width=\linewidth]{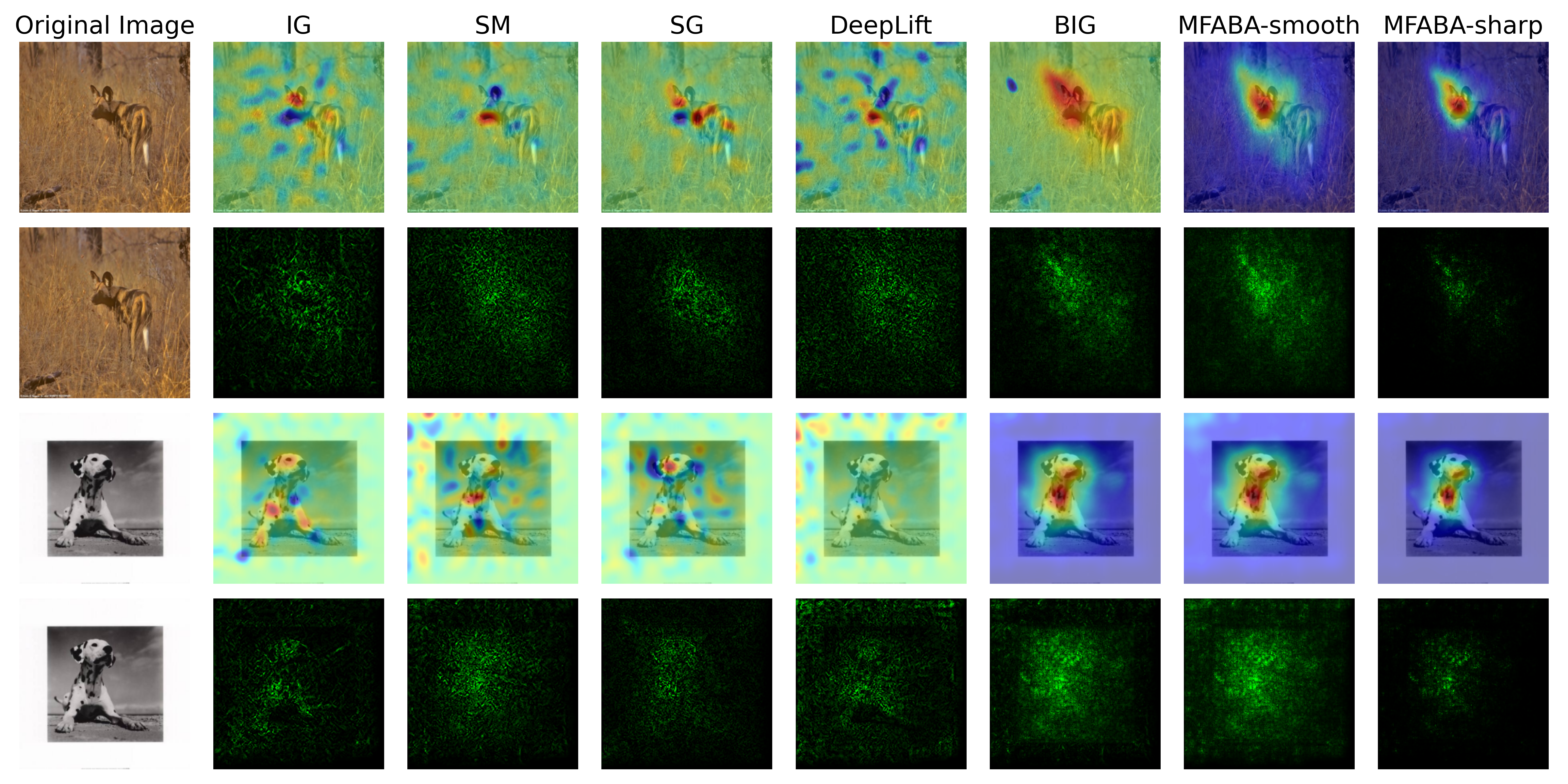}
    \end{subfigure}
    
    \caption{Results of MFABA compared to other SOTA methods (the colormaps demonstrate that our method can effectively highlight more concentrated regions associated with the recognized subjects, signifying higher interpretability)}
    \label{fig:comparison}
\end{figure}

        Figure.~\ref{fig:comparison} shows the results of the heatmap and the attribution map for different methods, including Integrated Gradients (IG), saliency map (SM), smoothed gradient (SG), DeepLift~\cite{shrikumar2017learning}, BIG~\cite{wang2021robust}, sharp and smooth gradient ascent methods based MFABA. It can be observed that, for MFABA results, the highlight areas are more focused related to the identified subjects, which can provide a better interpretation output. We have also provided more qualitative results in the Appendix folder.
 
	\subsection{Attribution Performance Evaluation}
	In addition to the visualised results for interpretation, herein we provide the statistical results for attribution performance evaluation, which are defined as the error rate evaluation indicators, the insertion and deletion score~\cite{petsiuk2018rise} and area under accuracy information curve~\cite{kapishnikov2019xrai}.
    \paragraph{Error rate evaluation indicators:}
    $$
Error Rate =\left|\frac{\sum a t t r\left(x^i\right)}{L\left(x_n\right)-L\left(x_0\right)}\right|
$$
	%\subsection{Error rate evaluation indicators}
   The error rate is obtained by dividing the final attribution result by the true attribution sum $L\left(x_n\right)-L\left(x_0\right)$. 
   We compare MFABA with its vanilla variant, which only utilises the first-order Taylor expansion for gradient information. We denote this method as `Vanilla' in Table~\ref{tab:error}. 
   %The vanilla version of MFABA employs a first-order Taylor expansion, while the advanced version is utilized by the second-order Taylor expansion. 
   We can see that the error rate of the MFABA is significantly reduced from Table~\ref{tab:error}. 
   Overall, the error rate is relatively low, which demonstrate the high efficiency of MFABA method. A detailed discussion can be found in the Appendix folder.
   \begin{table}[htbp]
\centering
\resizebox{0.35\textwidth}{!}{%
\begin{tabular}{c|c|c|c}
\hline
Dataset                   & Model                         & Method   & Error Rate \\ \hline
\multirow{2}{*}{ImageNet} & \multirow{2}{*}{EfficientNet} & Vanilla  & 0.02058    \\ \cline{3-4} 
                          &                               & MFABA & 0.01165    \\ \hline
\multirow{2}{*}{CIFAR10}  & \multirow{2}{*}{ResNet-50}     & Vanilla  & 0.2613     \\ \cline{3-4} 
                          &                               & MFABA & 0.01165    \\ \hline
\multirow{2}{*}{CIFAR100} & \multirow{2}{*}{ResNet-50}     & Vanilla  & 0.10221    \\ \cline{3-4} 
                          &                               & MFABA & 0.04192    \\ \hline
\end{tabular}%
}
\caption{Attribution Error Rate Results}
\label{tab:error}
\end{table}

\begin{figure}[htbp]
    \centering
    
    \begin{subfigure}{\linewidth}
        \centering
        \includegraphics[width=.9\linewidth]{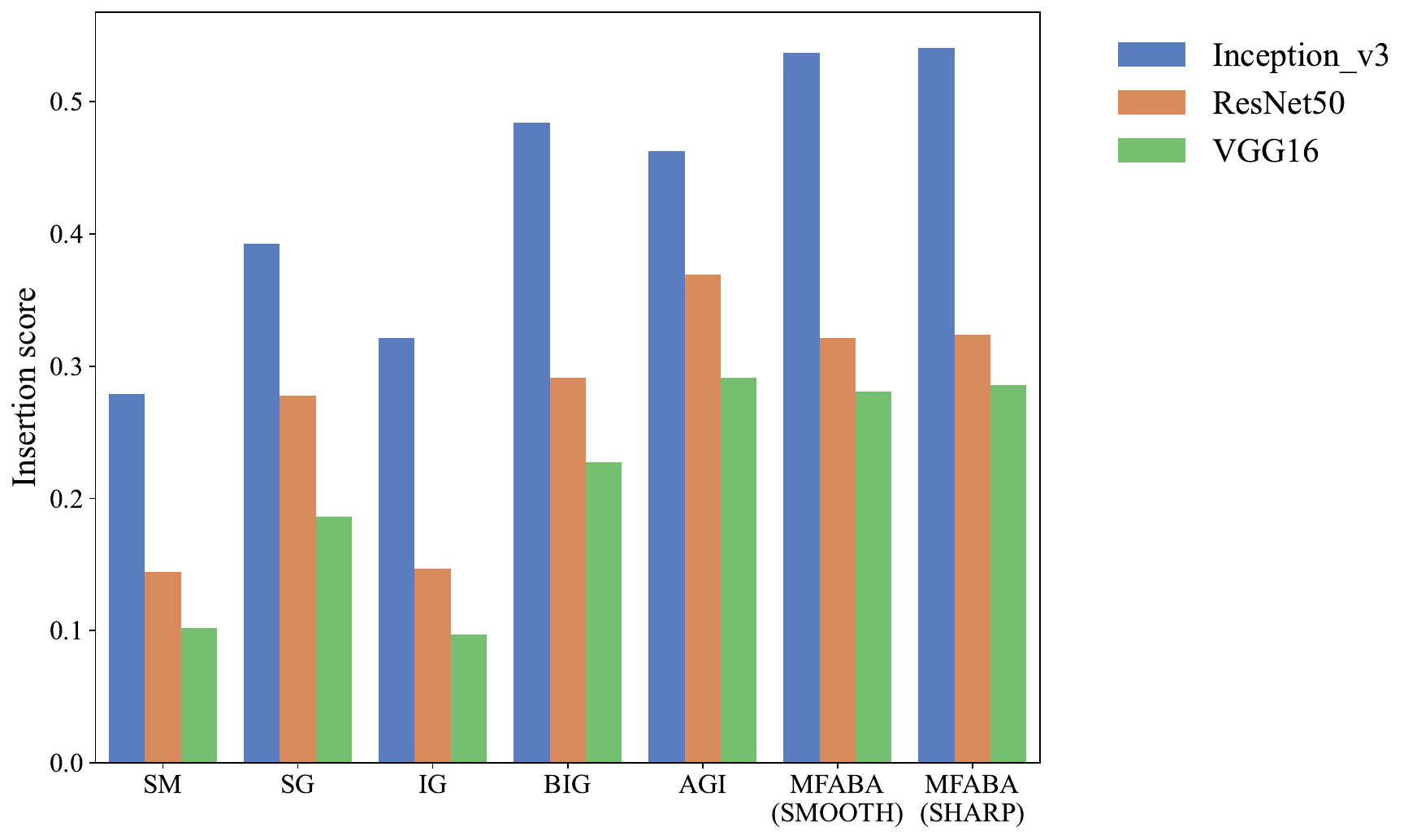}
        \label{subfig:empirical_evaluation}
    \end{subfigure}
    
    \begin{subfigure}{\linewidth}
        \centering
        \includegraphics[width=.9\linewidth]{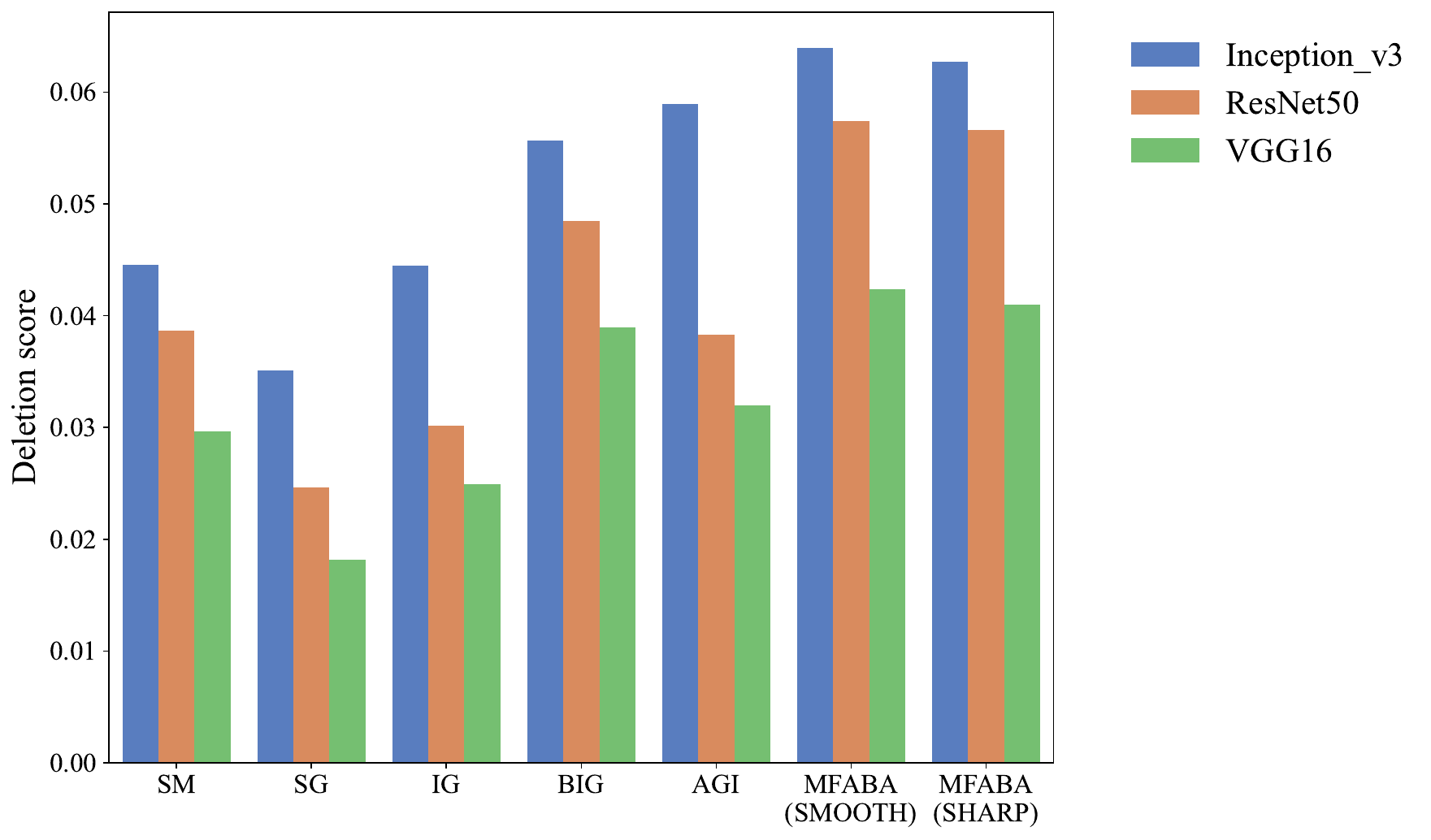}
    \end{subfigure}
    
    \caption{Insertion and Deletion score}
    \label{img:InsertandDelet}
\end{figure}

\paragraph{Insertion Score and Deletion Score}
Figure.~\ref{img:InsertandDelet} shows a schematic of our MFABA algorithm for the insertion and deletion scores. A higher insertion score corresponds to a more pronounced contribution of the input feature to the classification outcome. Conversely, a lower deletion score indicates an enhanced contribution of the input feature to the result. Our method exhibits a notable performance benefits on Inception-v3, closely trailing AGI on ResNet-50 and VGG-16. Since MFABA algorithm achieves a significant speed acceleration, we deem the trade-off to be both reasonable and acceptable.

\paragraph{Area Under Accuracy Information Curve}
In Table.~\ref{tab:Insert}, the employed evaluation criterion is the Area Under the Curve (AUC) of the Accuracy Information. This metric serves as an assessment tool to gauge the performance of interpretable algorithms with regard to the predictive accuracy of the model. It is evident that our method has achieved the most favorable outcomes among all the competing methods.

\begin{table}[htpb]
\resizebox{\columnwidth}{!}{%
\begin{tabular}{ccccc}
\hline
Model        & Method        & Insertion score & Deletion score & AUC    \\ \hline
Inception-v3 & SM            & 0.2792          & 0.0445         & 0.5150 \\
Inception-v3 & IG            & 0.3215          & 0.0445         & 0.5180 \\
Inception-v3 & BIG           & 0.4840          & 0.0557         & 0.5200 \\
Inception-v3 & AGI           & 0.4629          & 0.0590         & 0.5178 \\
Inception-v3 & MFABA(SMOOTH) & 0.5368          & 0.0640         & 0.5389 \\
Inception-v3 & MFABA(SHARP)  & 0.5407          & 0.0627         & 0.5367 \\ \hline
ResNet-50    & SM            & 0.1441          & 0.0387         & 0.4714 \\
ResNet-50    & IG            & 0.1467          & 0.0302         & 0.4823 \\
ResNet-50    & BIG           & 0.2911          & 0.0485         & 0.4759 \\
ResNet-50    & AGI           & 0.3695          & 0.0383         & 0.4772 \\
ResNet-50    & MFABA(SMOOTH) & 0.3211          & 0.0574         & 0.4854 \\
ResNet-50    & MFABA(SHARP)  & 0.3237          & 0.0566         & 0.4857 \\ \hline
VGG16        & SM            & 0.1018          & 0.0297         & 0.4257 \\
VGG16        & IG            & 0.0973          & 0.0249         & 0.4431 \\
VGG16        & BIG           & 0.2274          & 0.0390         & 0.4356 \\
VGG16        & AGI           & 0.2910          & 0.0320         & 0.4359 \\
VGG16        & MFABA(SMOOTH) & 0.2808          & 0.0424         & 0.4540 \\
VGG16        & MFABA(SHARP)  & 0.2856          & 0.0410         & 0.4540 \\ \hline
\end{tabular}
}
\caption{Insertion score (the higher the better), deletion score (the lower the better), and AUC (the higher the better)}
\label{tab:Insert}
\end{table}

\subsection{FPS Results}
	We use FPS, which refers to the number of frames per second (FPS) processed by the algorithms, to evaluate the algorithm processing speed. The hardware for our experiment includes: RTX 3090(24GB)*1 for GPU, 24 vCPU AMD EPYC 7642 48-Core for CPU and 80GB RAM. To comprehensively evaluate the algorithm efficiency, we have conducted two separate tests including single image testing and multiple images testing.
 
    The single image test is the calculation time for one image at a time. An average result is obtained against the experiment datasets. For the multiple images testing, we count the maximum number of images being processed per second at the same time with the same specified hardware conditions. We have run the experiment for three times, and the average value is returned. We observed that, MFABA requires a smaller shared memory in GPU environment. Thus, with MFABA, we are able to deploy a larger batch size of images for attribution computation.

\begin{table}[htpb]
\centering
\resizebox{0.45\textwidth}{!}{%
\begin{tabular}{@{}c|c|c|c|c|c@{}}
\toprule
Dataset  & Method       & BIG    & IG      & AGI    & MFABA    \\ \midrule
CIFAR10  & ResNet-50     & 1.3492 & 24.6376 & 0.1404 & 136.9630 \\ \midrule
CIFAR100 & ResNet-50     & 1.36   & 24.2857 & 0.1489 & 162.7396 \\ \midrule
ImageNet & ResNet-50     & 0.3704 & 6.7568  & 0.2846 & 51.5827  \\ \midrule
ImageNet & EfficientNet & 0.3205 & 10.4167 & 0.2096 & 39.9685  \\ \bottomrule
\end{tabular}%
}
\caption{FPS Results of BIG, IG, AGI, MFABA Algorithms}
\label{tab:fps}
\end{table}

     In Table~\ref{tab:fps}, the results for multiple images testing is presented. MFABA has demonstrated its superiority over the other state-of-the-art methods, achieving a speed increase of more than 101 times compared to BIG (for CIFAR10 dataset with ResNet-50 model), and up to 139.2621 times faster than BIG (for ImageNet dataset with ResNet-50). In comparison with IG (for ImageNet with EfficientNet), MFABA achieves near 4 times speed increase, and near 8 times faster for ImageNet dataset with ResNet-50. For AGI method, MFABA is at least 181.2463 times faster for ImageNet with ResNet-50, and over 1000 times for CIFAR100 with ResNet-50. Overall, MFABA has achieved the best performance in all categories of the experiments with the datasets of CIFAR10, CIFAR100 and ImageNet.

\section{Conclusion}
In this paper, we present MFABA algorithm, a more faithful and accelerated attribution algorithm for deep neural networks interpretation. Particularly, we have proposed two distinct versions with sharp and smooth methods for MFABA. We also provide the proof for the axiomatic derivation process for MFABA, which supports the two fundamental axioms of Sensitivity and Implementation Invariance. A large scale experiment is conducted, demonstrating the state-of-the-art performance of MFABA. In addition, we provide in-depth analysis with experiment evidence for the performance of MFABA, such as we argue that aggressive samples can substantially contribute to the attribution output. The complete replication package is open-sourced, and we hope it will contribute to future research in advancing trustworthy AI in this field. It is important to note, however, that our evaluation is currently limited to the conventional image dataset, omitting more intricate image tasks. Future efforts will encompass the application of MFABA in a broader range of scenarios to comprehensively assess the performance.

\bibliography{aaai24}

\newpage
\appendix
\onecolumn
\section{A \quad Appendix}
In this supplementary material, we have conducted the experiments to support our analysis in Section. \textbf{In-depth analysis} and Section. \textbf{Evaluation}. We show the details about the specific performance and relevant schematic diagrams of our MFABA algorithm under different conditions.

\subsection{A.1 \quad Impact of non-aggressive samples on MFABA algorithm}

\subsection{A.1.1 \quad Heat map analysis of figure 2 in our paper}
In Figure. 2 of the main paper, we have visualized the results with and without the non-aggressive sample for the linear gradient ascending path in MFABA. We can see that, the left diagram is better than right one, supporting our analysis in Section. \textbf{Definition of Aggressiveness}.

\subsection{A.1.2 \quad Additional figures for Comparison of heatmap with and without non-aggressive samples}
    	\begin{figure}[H] 
		\centering
		\centerline{\includegraphics[width=0.9\linewidth]{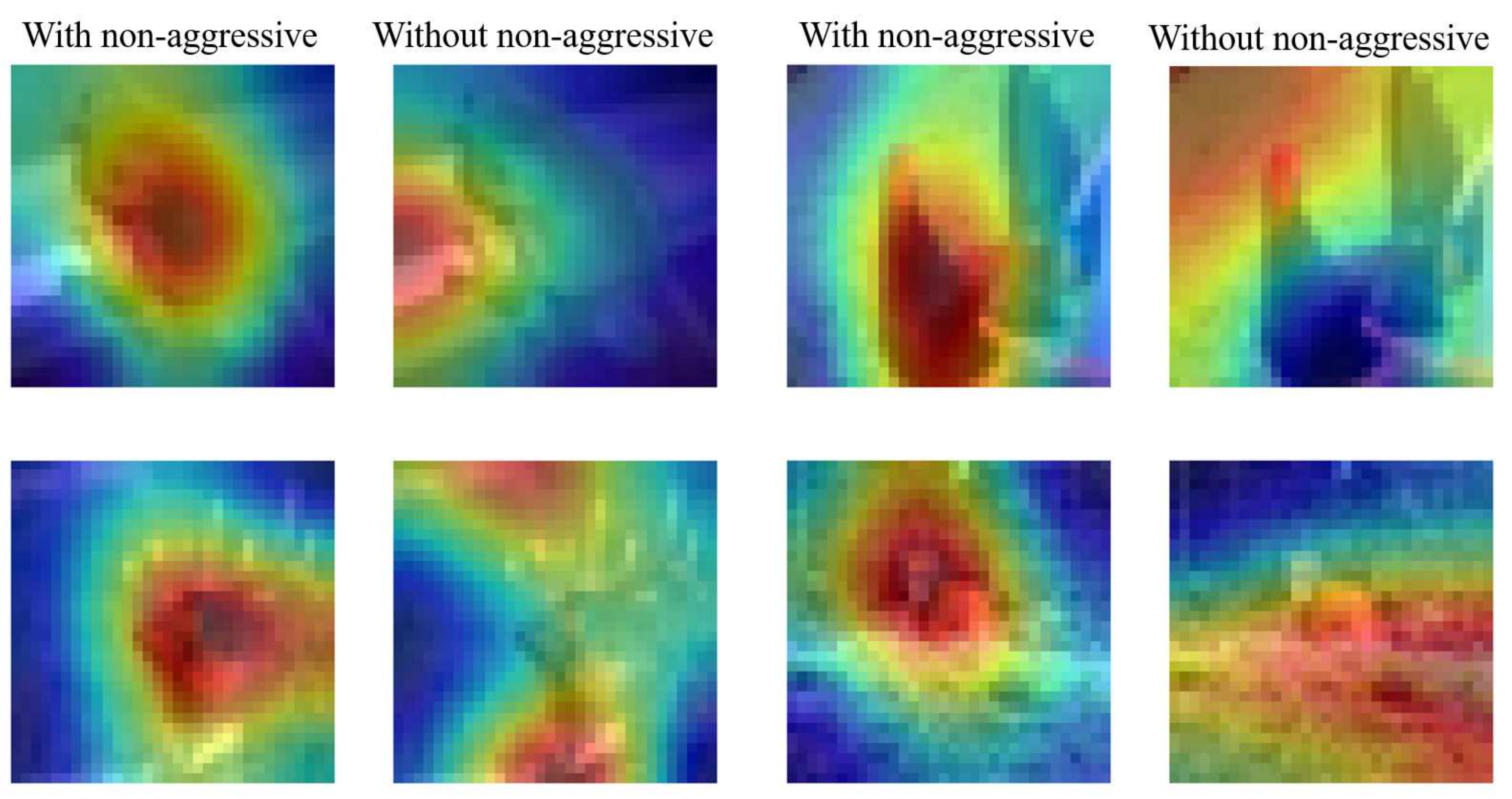}}
		%\caption{}
		\centering
	\end{figure}

\subsection{A.2 \quad Comparison of MFABA-cosine and MFABA-norm}

\subsection{A.2.1 \quad Results for BIG Algorithm, and MFABA Approximation Algorithms about Eq. 18 and Eq. 19}

\begin{figure}[t]
		%\vskip 0.2in
		\begin{center}
			\centerline{\includegraphics[scale=0.3]{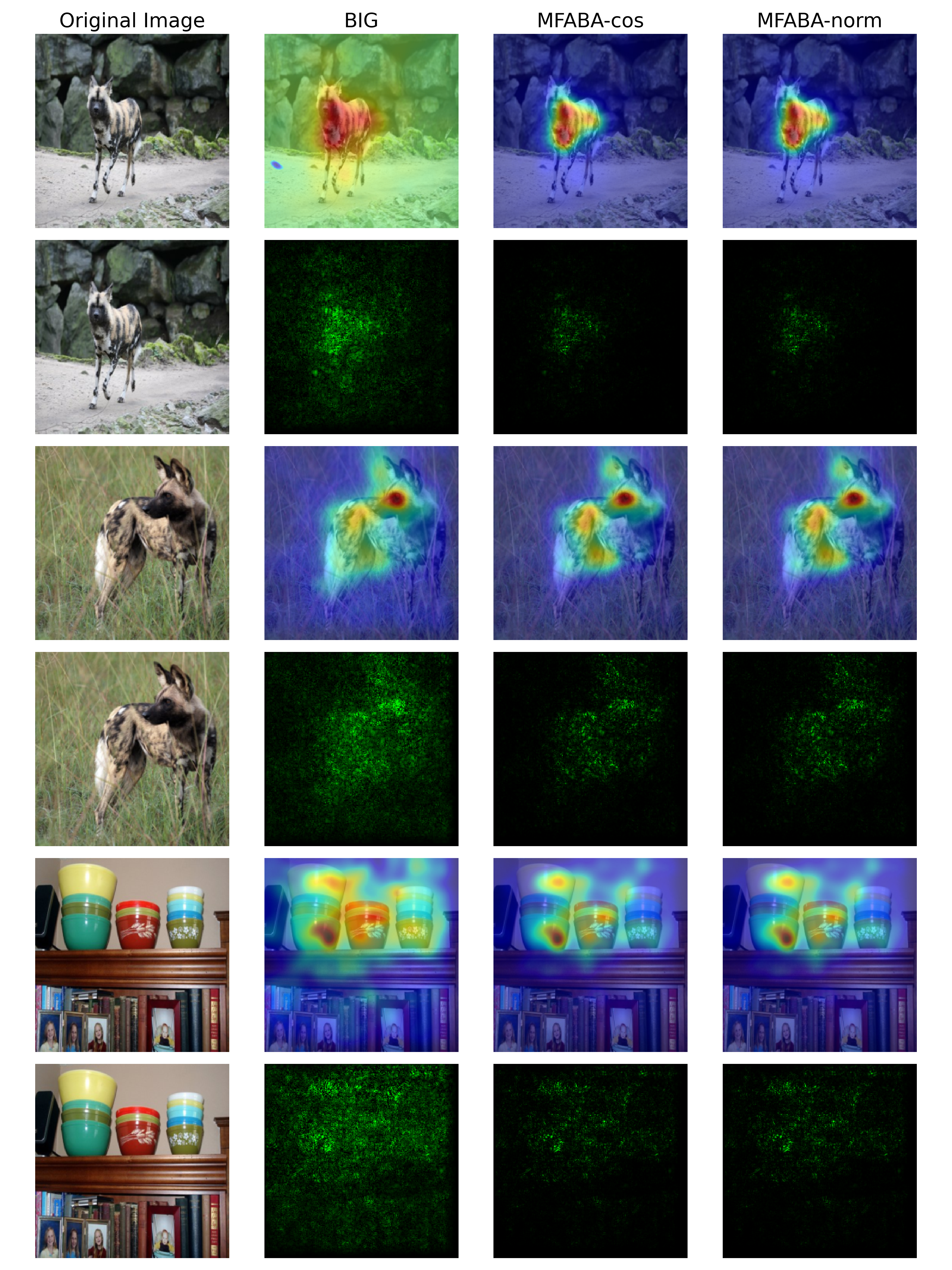}}
			\caption{Results for BIG Algorithm, and MFABA Approximation Algorithms about Eq. 18 and Eq. 19}
			\label{img:ablation_approximation}
		\end{center}
\end{figure}

In Figure.~\ref{img:ablation_approximation}, we compare the results of BIG, MFABA with cos method and MFABA with norm method separately. For the cos and norm methods, they can be referred to Eq. 18 and 19. Figure~\ref{img:ablation_approximation} demonstrates the linear MFABA is able to approximate BIG algorithm and has less noise in the attribution output, supporting our analysis in Section. \textbf{Comparision with Other State-fo-the-art Methods}.

\subsection{A.2.2 \quad Additional figures for BIG Algorithm, and MFABA Approximation Algorithms about Eq. 18 and Eq. 19}

    	\begin{figure}[H] 
		\centering
		\centerline{\includegraphics[width=0.8\linewidth]{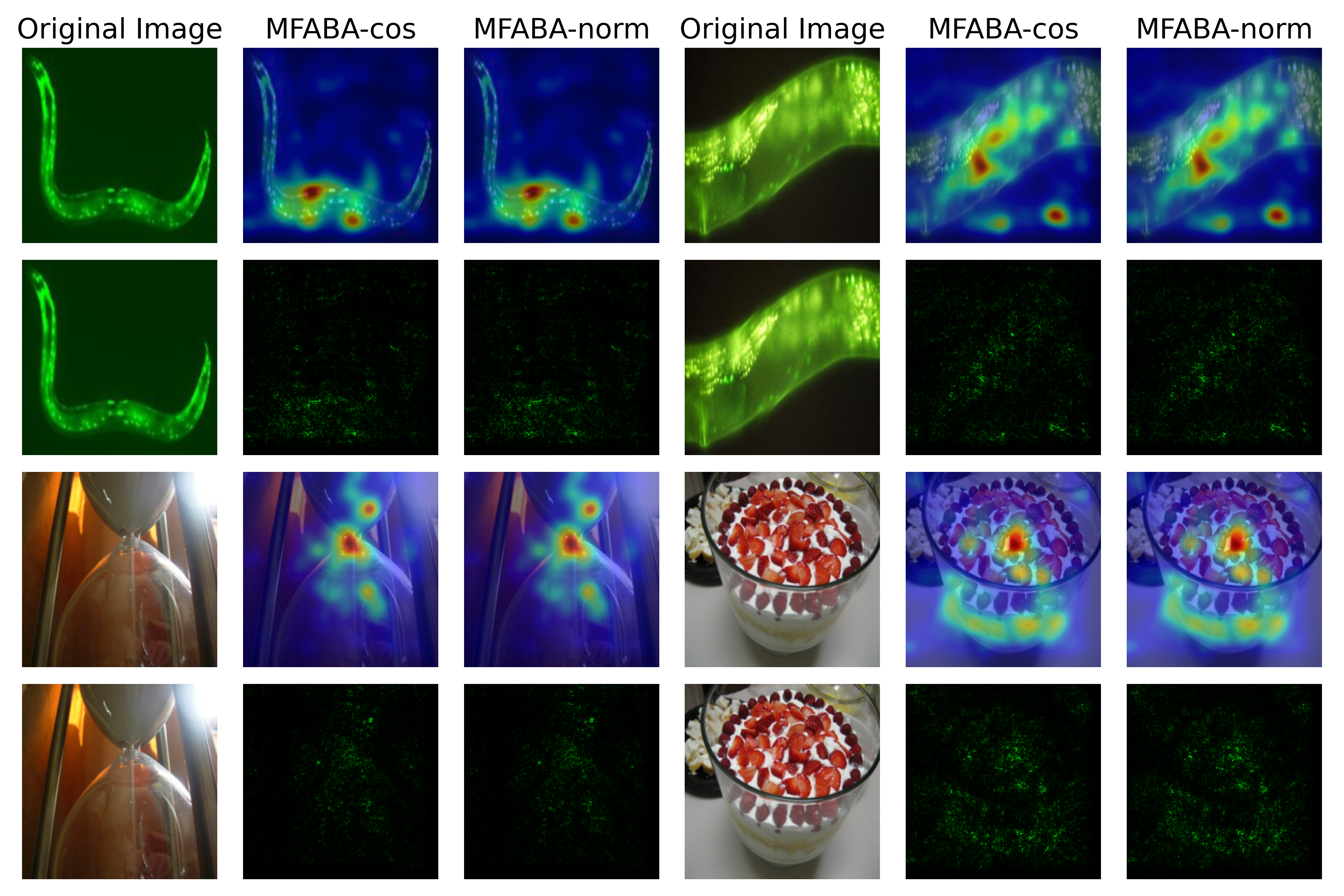}}
		%\caption{}
		\centering
	\end{figure}

\subsection{A.3 \quad Evaluation of MFABA on Imagenet Dataset}

\subsection{A.3.1 \quad Evaluation of MFABA with and without the softmax on Imagenet Dataset}
\begin{figure}[htbp]
		%\vskip 0.2in
		%\begin{center}
		\centerline{\includegraphics[width=0.65\linewidth]{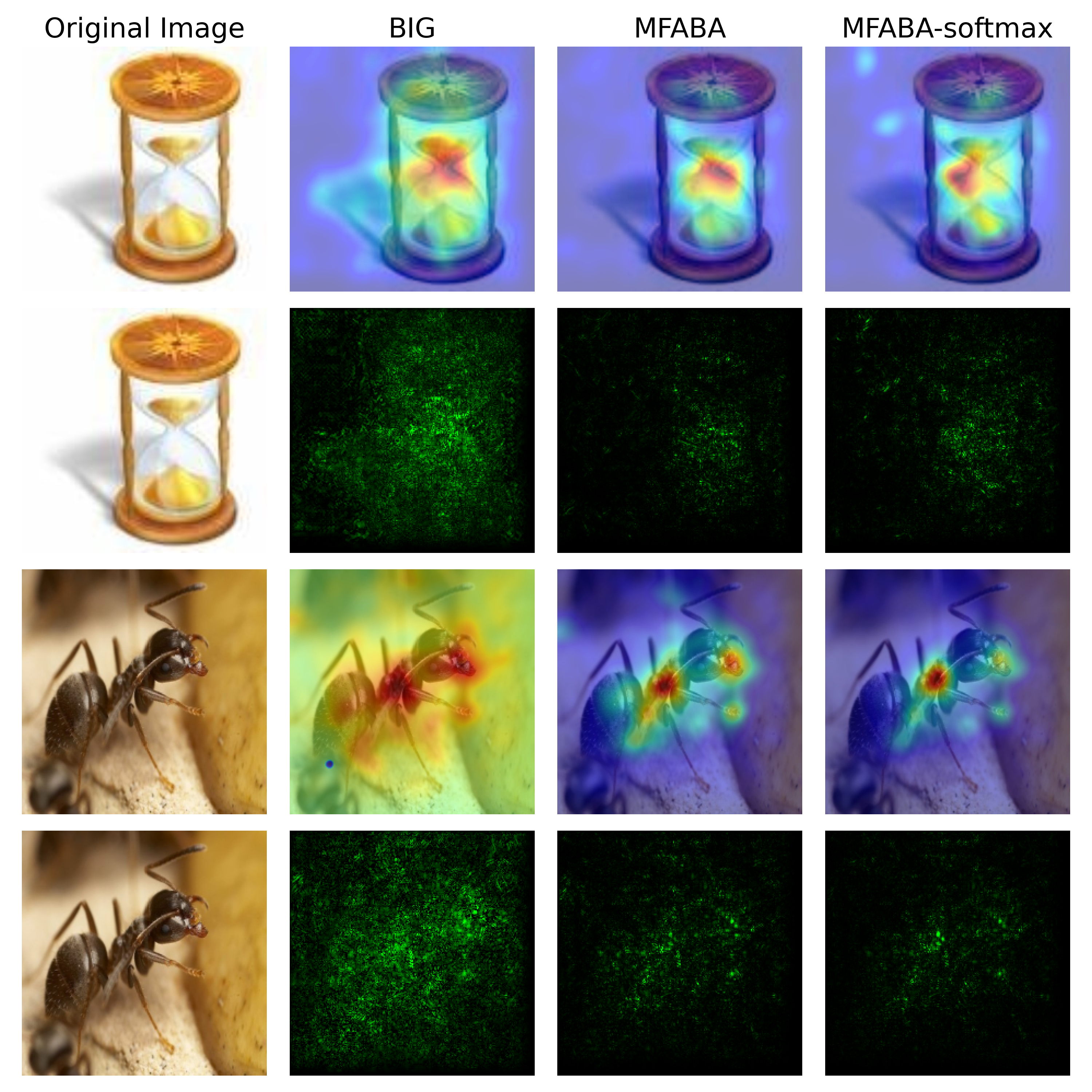}}
		\caption{Evaluation of MFABA with and without the softmax for Imagenet Dataset}
		\label{img:ablation_softmax}
		%\end{center}
\end{figure}

In Figure~\ref{img:ablation_softmax}, we can see that the fitness of softmax is higher, which provide better visualization results and support our analysis in Section. \textbf{Attribution Method in MFABA}. 

% \newpage
\subsection{A.3.2 \quad Additional figures for Evaluation Results of MFABA with and without Softmax for Imagenet Dataset}

\begin{figure}[htbp]
	\centering
	\centerline{\includegraphics[width=0.9\linewidth]{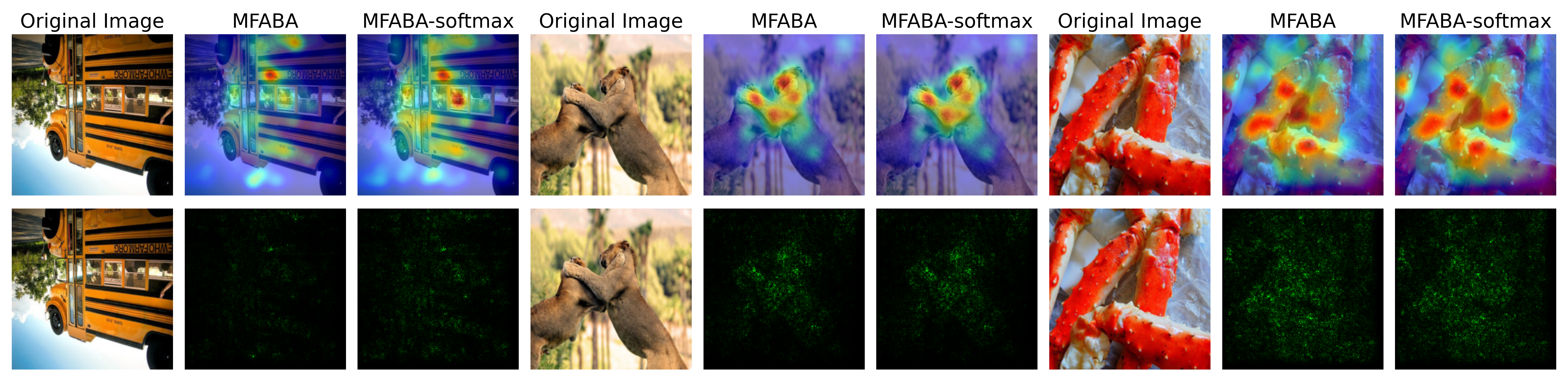}}
		%\caption{}
	\centering
\end{figure} 

\subsection{A.4 \quad Additional figures for attribution results of MFABA in comparison with other state-of-the-art methods}

In this section we provide more quantitative visualization results (the figure below and figure. 3), supporting our analysis in Section. \textbf{Empirical Evaluation}.

\begin{figure}[hbbp]
	\centering
	\centerline{\includegraphics[width=\linewidth]{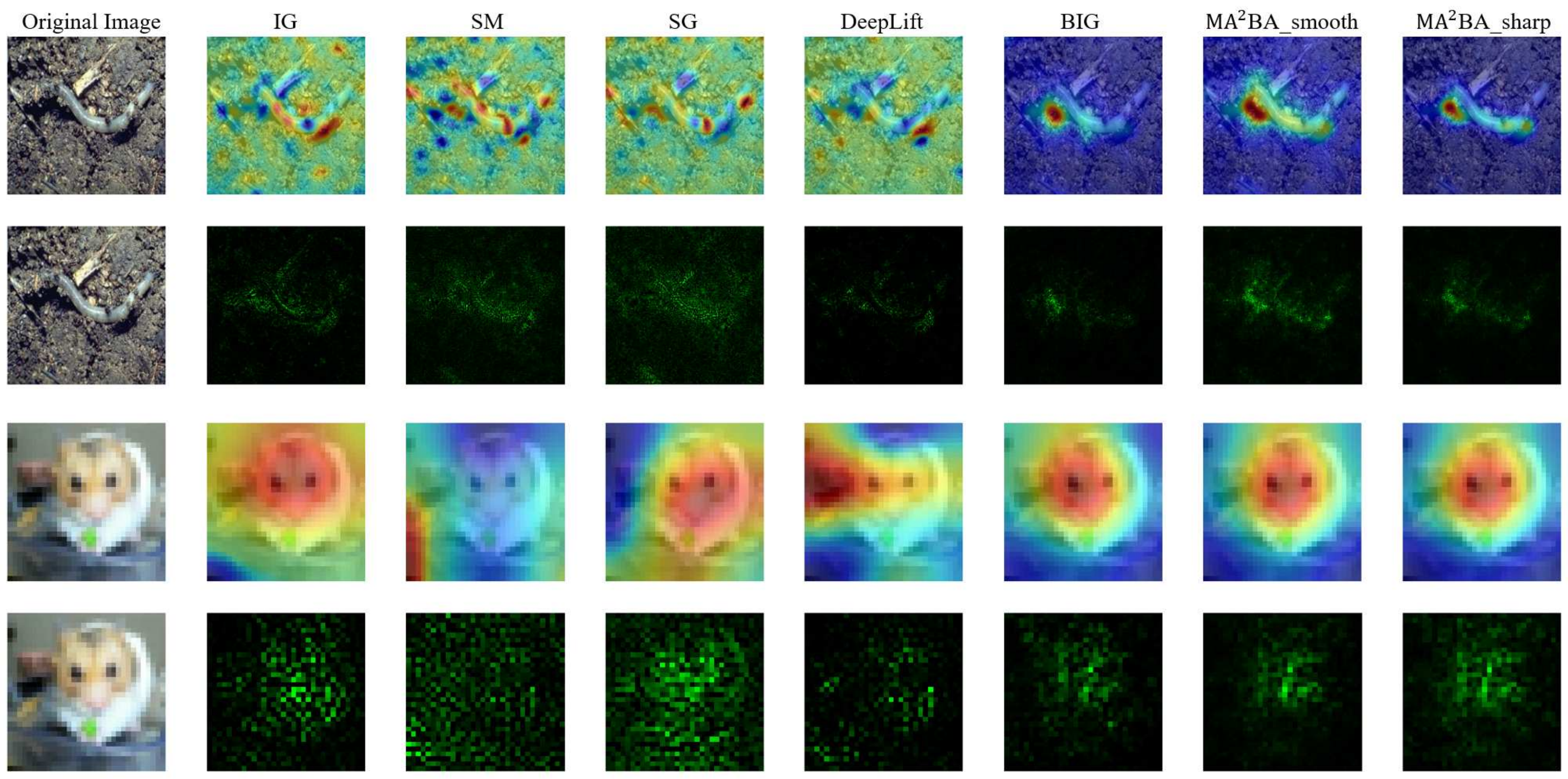}}
		%\caption{Results of MFABA compared to other methods}
	\centering
\end{figure}

\begin{figure}[hbbp]
	\centering
	\centerline{\includegraphics[width=\linewidth]{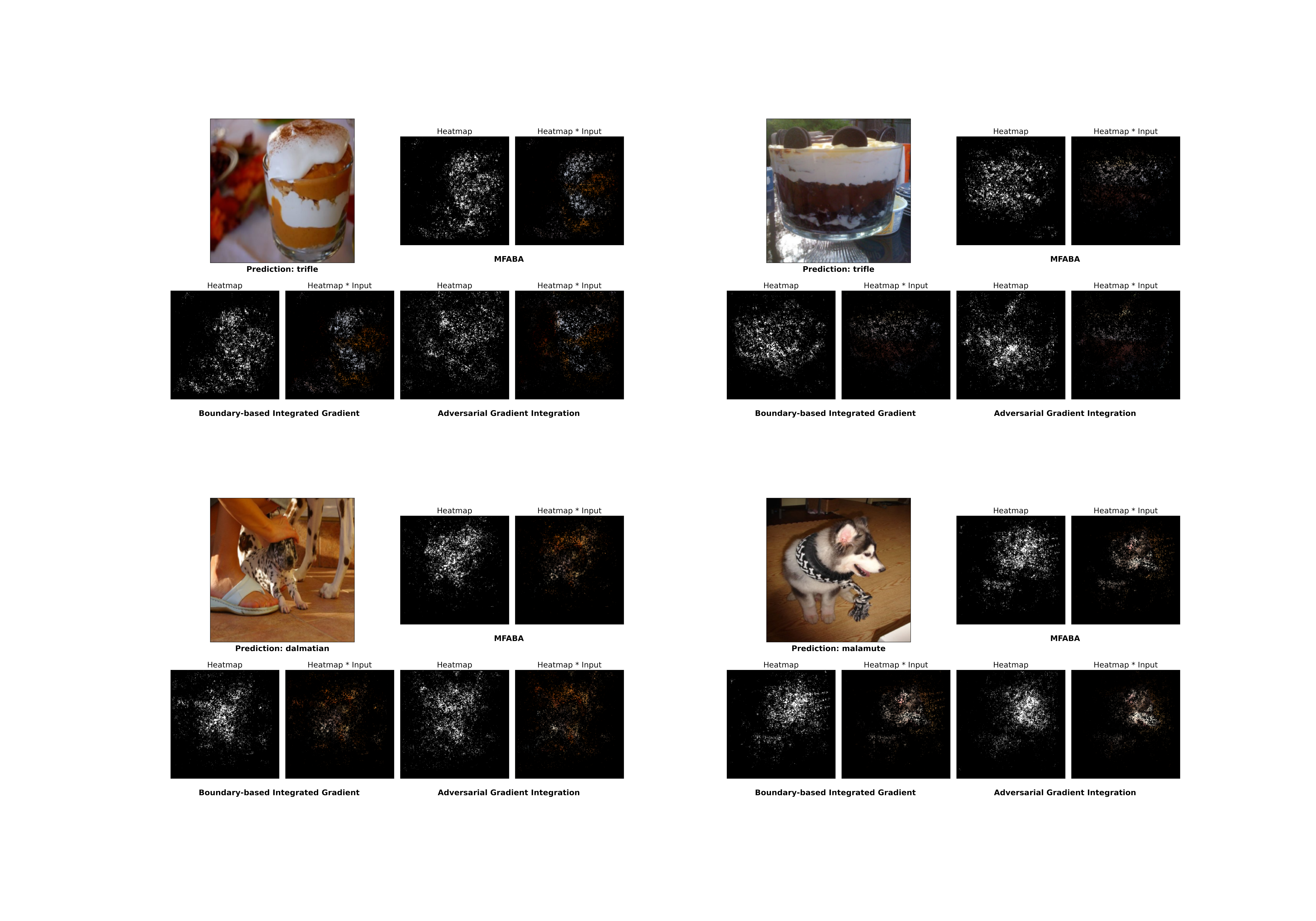}}
	\caption{Additional figures for attribution results of MFABA in comparison with other state-of-the-art methods}
	\centering
\end{figure}

\subsection{A.5 \quad Algorithm Procedure}
 As shown in the following pseudo code, the gradient ascent method is used to line up adversarial data pairs, and each gradient ascent will return the corresponding gradient degree $\frac{\partial L\left(x_j\right)}{\partial x_j}$, the attack result $x_j$ and the incoming label $y'$. $y'$ is the new label for the sample to determine whether the gradient ascent will be stopped. The attribution $\sum_{j=0}^{n-1} \frac{\left(\frac{\partial L\left(x_j\right)}{\partial x_j}+\frac{\partial L\left(x_{j+1}\right)}{\partial x_{j+1}}\right)}{2}\left(x_{j+1}^i-x_j^i\right)$ is then given according to Algorithm 1, which optimises the summation rate by computing dot product and finally returns the corresponding attribution result.

	\begin{algorithm}[h]
		\label{alg:1}
		\caption{MFABA($f$,$m$,$x_0$,$y$)}
		%\begin{algorithmic}
		{\bfseries Input:} model $m$,target $f$,input $x_0$,label $y$,Iterative number $n$,learning rate $\eta$,method
		
		$X=[x_0]$, $grads=[]$, $j=1$
		
		\textbf{for}	$j \leq n${\bfseries}	\textbf{do}	
		
		\qquad$grad=\frac{\partial L\left(x_{j-1}\right)}{\partial x_{j-1}}$
		
		\qquad $x_j=x_j+method(grad)$
		
		\qquad$x_j \in R^{w\times h \times 3}grad \in R^{w\times h \times 3}$
		
		\qquad$x.append(x_j)$
		
		\qquad$y'=m(x)$
		
		\qquad$grads.append(grad)$
		
		\qquad$j=j+1$
		
		\qquad\textbf{if}  $y' \neq y$
		
		\qquad\qquad\textbf{break}
		
		\textbf{end for}
		
		$grads.append(\frac{\partial f\left(x_{j}\right)}{\partial x_{j}})$
		
		$addr=-(x[1:]-x[:-1])\cdot \frac{(\operatorname{grads}[:-1]+\operatorname{grads}[1:])}2$
		
		\textbf{return} addr
	\end{algorithm}

\end{document}